\documentclass{article}
\usepackage{iclr2027_conference,times}

\usepackage{amsmath,amsfonts,bm}

\def\Figref#1{Figure~\ref{#1}}

\def\Secref#1{Section~\ref{#1}}

\def\eqref#1{equation~\ref{#1}}

\def\Algref#1{Algorithm~\ref{#1}}

\def\1{\bm{1}}

\def\rvepsilon{{\mathbf{\epsilon}}}

\def\rva{{\mathbf{a}}}

\def\vzero{{\bm{0}}}

\def\vmu{{\bm{\mu}}}
\def\vtheta{{\bm{\theta}}}
\def\va{{\bm{a}}}

\def\vc{{\bm{c}}}

\def\ve{{\bm{e}}}

\def\vh{{\bm{h}}}

\def\vo{{\bm{o}}}

\def\vs{{\bm{s}}}

\def\vv{{\bm{v}}}

\def\vx{{\bm{x}}}

\def\vz{{\bm{z}}}

\def\mI{{\bm{I}}}

\def\mQ{{\bm{Q}}}

\def\mSigma{{\bm{\Sigma}}}

\DeclareMathAlphabet{\mathsfit}{\encodingdefault}{\sfdefault}{m}{sl}
\SetMathAlphabet{\mathsfit}{bold}{\encodingdefault}{\sfdefault}{bx}{n}

\newcommand{\E}{\mathbb{E}}

\newcommand{\KL}{D_{\mathrm{KL}}}

\usepackage{hyperref}
\usepackage{url}
\usepackage{fontawesome5}
\usepackage{graphicx}
\usepackage{placeins}
\usepackage{booktabs}
\usepackage[table]{xcolor}
 \usepackage{multirow}
 \usepackage{amssymb}
 \usepackage{tabularx}
 \usepackage{pifont}
\usepackage{subcaption}
\newcommand{\cmark}{\ding{51}}
\newcommand{\xmark}{\ding{55}}
\newcommand{\Appref}[1]{Appendix~\ref{#1}}
\newcommand{\Tabref}[1]{Table~\ref{#1}}
\usepackage{array}
\usepackage{algorithm}
\usepackage{algpseudocode}

\title{\mbox{\raisebox{-0.18\height}{\includegraphics[height=1.15em]{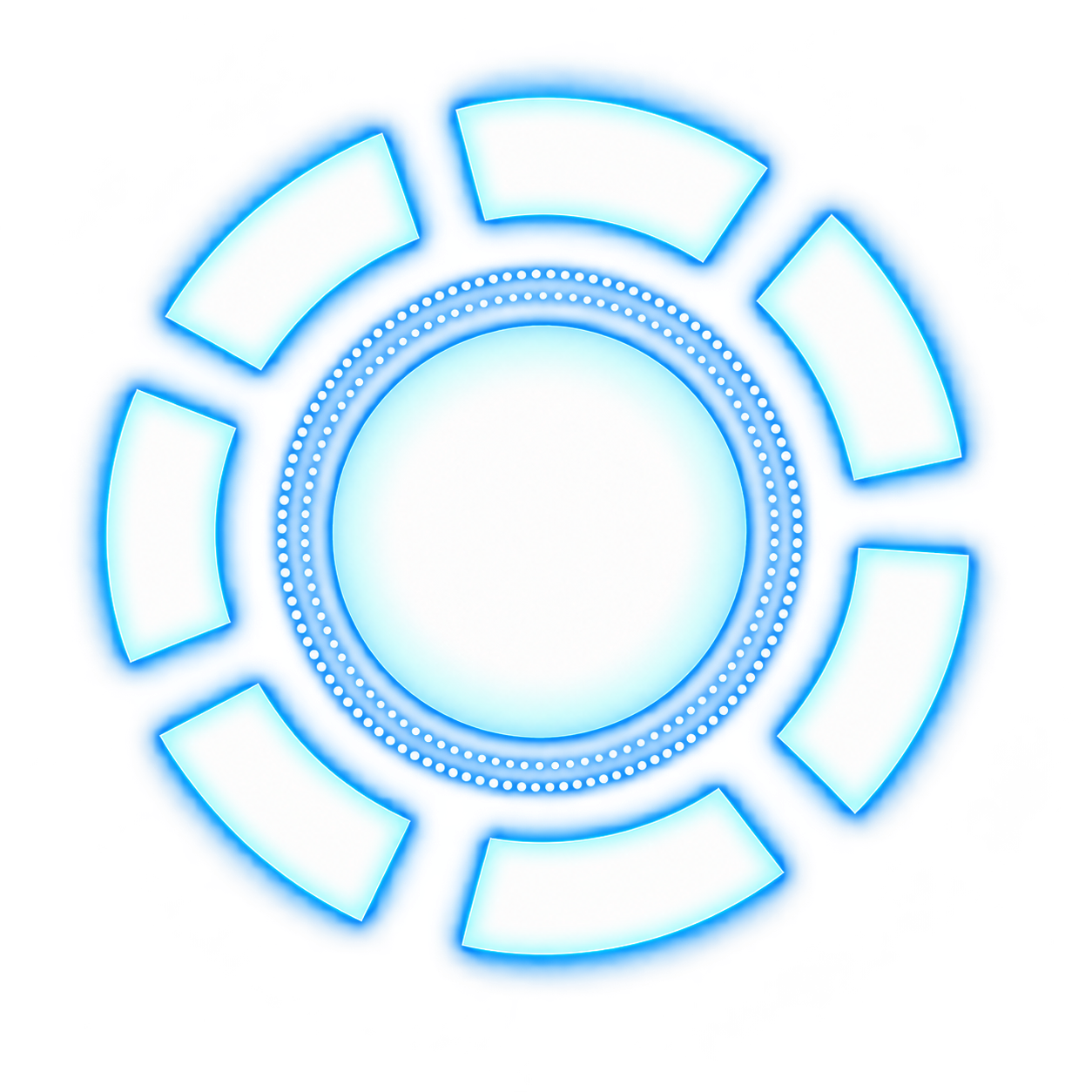}}\,IronMan: Information-Constrained Video-Action}\\
Learning for Robot Manipulation}

\author{Yuanshuo Zhang\textsuperscript{1} \quad Wenzhe Zhao\textsuperscript{2} \quad Zixing Lei\textsuperscript{1} \quad Bin Chen\textsuperscript{2} \quad Siheng Chen\textsuperscript{1}\thanks{Corresponding author.} \\
\textsuperscript{1}Shanghai Jiao Tong University \quad \textsuperscript{2}Joy Future Academy, JD
}

\iclrfinalcopy
\begin{document}

\maketitle
\begin{abstract}
Video Action Models~(VAMs) couple visual dynamics modeling with action generation for robot manipulation. However, video representations are not naturally suited to action generation, as exposing the action policy to excessive visual detail can impair its generalization ability. Therefore, we introduce \textbf{IronMan} (\textbf{I}nformation-const\textbf{R}ained vide\textbf{O}-actio\textbf{N} learning for robot \textbf{MAN}ipulation), a robust video-action learning framework built on information bottleneck principle. The core principle of this framework is to impose information constraints that suppress irrelevant visual information while preserving action-relevant dynamics cues. IronMan employs a dynamics-aware bottleneck that distills noisy, entangled one-step video features into compact world representations. Extensive simulation and real-world experiments demonstrate strong in-distribution~(ID) performance and out-of-distribution~(OOD) robustness while maintaining efficient inference.  IronMan achieves success rates of 99.0\% on LIBERO and 79.4\% on RoboTwin clean2clean, outperforming all the evaluated baselines. Under OOD shifts, IronMan achieves a success rate of 79.1\% on LIBERO-Plus, exceeding the strongest baseline by 10.4 percentage points. Project Page: \url{https://youngsoul0731.github.io/ironman-project-page/}

\end{abstract}

\section{Introduction}
\label{sec:introduction}

Vision-Language-Action (VLA) models have emerged as a prominent class of foundation models for robot manipulation~\citep{kim2024openvla,black2024pi0,physicalintelligence2025pi05,intelligence2025pi06vlalearnsexperience,intelligence2026pi07steerablegeneralistrobotic,wu2026lingbotvla,liu2026g05autoregressivestreamrobot}. However, most VLAs inherit priors from Vision-Language Models (VLMs) pretrained primarily on static image--text data, and therefore lack dynamics modeling.
This limitation has motivated a video-action learning paradigm that couples visual dynamics with action generation, enabling these Video-Action Models~(VAMs) to learn inverse dynamics within predictive representations~\citep{ICLR2025_e5b5c402,hu2025vpp,bi2026motus,ye2026dreamzero,li2026lingbotva,yuan2026fastwam,nvidia2025cosmosworldfoundationmodel,zhang2026disentangled}. These advances reveal the great
potential of jointly modeling video dynamics and actions.

However, video and action generation inherently impose different information requirements. A video backbone needs to preserve rich visual details to reconstruct future frames. But an action policy should primarily leverages action-relevant cues. Therefore, the rich information from video backbones is not naturally aligned with the needs of action generation. Existing approaches typically connect video features to the action policy through learned projections or attention mechanisms~\citep{yuan2026fastwam,hu2025vpp,yan2026svam}. Such interfaces do not explicitly suppress features irrelevant to action generation. 

This can undermine the robustness of action policy. When excessive visual cues are exposed to the policy, the model may exploit incidental correlations, which are unreliable under visual distribution shifts. For example, camera noise, light conditions, and background may become shortcuts, leading to degraded generalization in unseen scenes. Therefore, a key challenge is how to transform rich video features into compact representations that capture the dynamics essential for manipulation while avoiding spurious visual correlations.

The information bottleneck~(IB) theory offers a principled way to regulate the video-action interface by constraining how much information can pass from input features, while encouraging the latent representation to retain sufficient information for prediction target~\citep{tishby1999information,alemi2017deep}. Prior work applies IB-style objectives to regularize visuomotor representations for robot manipulation~\citep{bai2025rethinking,zhang2026information,pmlr-v162-du22b}. However, these methods do not address the more challenging video-action interface, where features from a generative video backbone entangle predictive dynamics with visual details.  

\begin{figure}[ht]
    \centering
    \includegraphics[width=\linewidth]{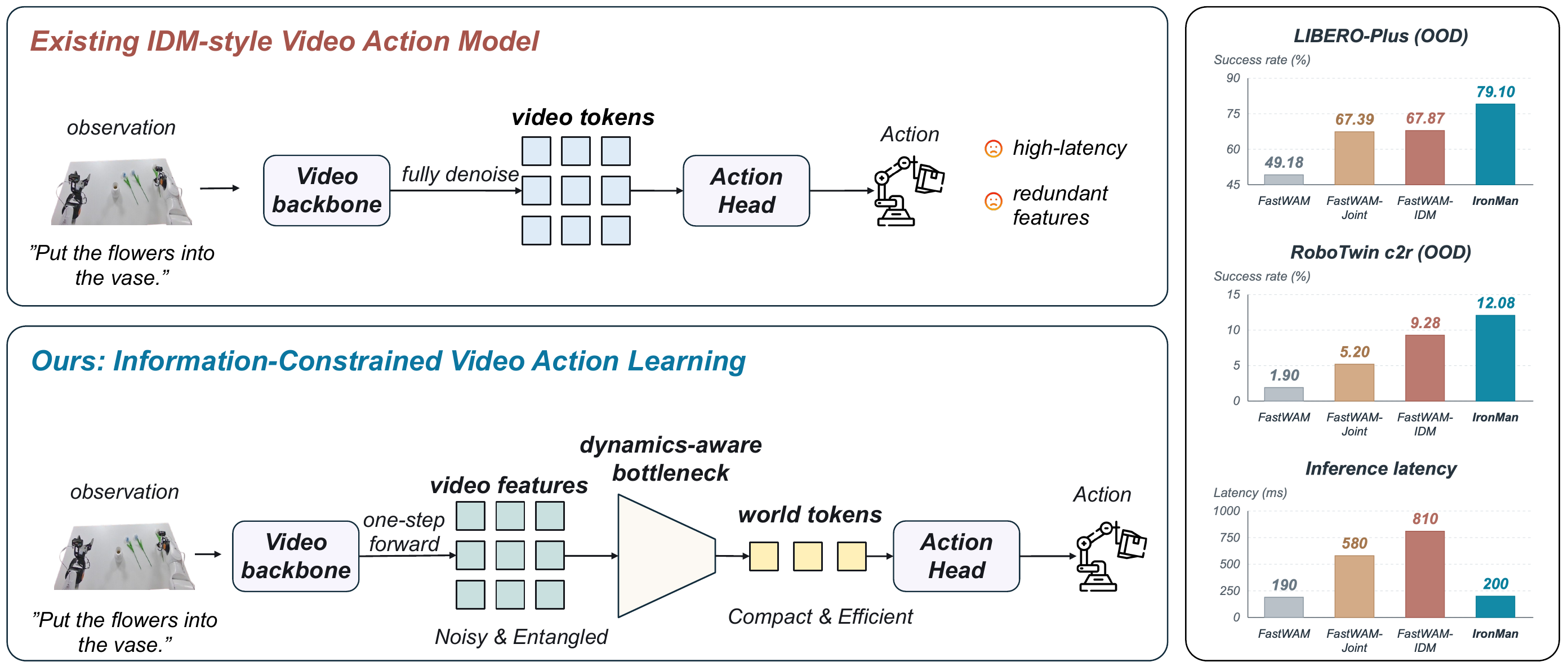}
    \caption{Motivation of IronMan. Existing video--action interfaces allow action policy to freely attend to irrelevant visual signals, increasing its sensitivity to visual details. IronMan instead uses an information-constrained bottleneck designed to retain the predictive dynamics needed for action while suppressing redundancy.}
    \label{fig:motivation}
\end{figure}

Motivated by this, we introduce \textbf{IronMan} (\textbf{I}nformation-const\textbf{R}ained vide\textbf{O}-actio\textbf{N} learning for robot \textbf{MAN}ipulation), a robust video-action learning framework. As illustrated in \Figref{fig:motivation}, unlike inverse dynamics model (IDM)-based VAMs, IronMan extracts one-step video features from the video backbone, substantially reducing inference overhead. Although these features already encode physical dynamics, they remain highly noisy and entangled with redundant visual cues. Applying an information constraint on these features alone does not specify how spatial and temporal cues should be used for action generation. Therefore, we introduce a \textbf{dynamics-aware bottleneck} in~\Figref{fig:model_arch} which first aggregates the video features with a Qformer-style module~\citep{li2023blip}, and then maps the aggregated features to a parameterized latent distribution through a Gaussian head. These latents constitute world tokens, forming an efficient and compact video-action interface. To instantiate the information bottleneck in our video-action learning framework, we adopt a variational formulation~\citep{alemi2017deep} that penalizes the KL divergence between the latent distribution and a standard Gaussian prior, and combine this regularization with video and action supervision to form our joint learning objective.

To evaluate the effectiveness of IronMan, we conduct simulation experiments across two datasets and four benchmarks covering both in-distribution (ID) and out-of-distribution (OOD) settings: LIBERO, LIBERO-Plus, RoboTwin clean2clean, and RoboTwin clean2random. We further evaluate three real-world manipulation tasks under ID and OOD settings with background and layout shifts. Across these evaluations, IronMan achieves strong ID performance and OOD robustness while maintaining efficient inference.

\section{Related Work}
\label{sec:realated}
\vspace{-1em}
\subsection{Vision Language Action Models}
\vspace{-0.5em}
Vision--Language--Action (VLA) models, such as RT-2~\citep{brohan2023rt2}, OpenVLA~\citep{kim2024openvla}, $\pi_0$~\citep{black2024pi0}, $\pi_{0.5}$~\citep{physicalintelligence2025pi05}, and GR00T~\citep{nvidia2025gr00t}, have emerged as a powerful paradigm for robot manipulation by transferring the rich visual and semantic knowledge of pretrained vision--language models (VLMs) to embodied control. Leveraging large-scale pretraining on diverse image--text data, VLAs enable robots to better understand instructions, recognize a wide range of objects and scenes, and generalize across tasks and environments beyond the limited coverage of robotic datasets. This has substantially improved the versatility and generalization of robotic policies, establishing VLAs as a prominent direction toward general-purpose robot manipulation. Despite these achievements, modeling physical dynamics remains a fundamental limitation of current VLAs. Because their VLM backbones are pretrained predominantly on static image--text pairs, they receive little direct supervision about temporal evolution. Recent approaches seek to mitigate this limitation during post-training by aligning VLM representations with features of future observations extracted by vision foundation models, thereby injecting dynamics into a VLM backbone~\citep{zheng2025flare,su2026wog,xu2026futurevla,luo2026beingh07latentworldactionmodel,cen2025WorldVLA,cen2025rynnvla}. In contrast, our work directly uses a pretrained video generative model as the source of physical dynamics priors.

\vspace{-1em}
\subsection{Video Action Models}
\vspace{-0.5em}
Recent Video-Action Models~(VAMs) increasingly leverage generative video models to provide rich physical dynamics priors~\citep{ICLR2025_e5b5c402,hu2025vpp,bi2026motus,ye2026dreamzero,li2026lingbotva,yuan2026fastwam}. A paradigm is explicit future prediction, where future frames are generated to guide action generation~\citep{li2026lingbotva,ye2026dreamzero, bharadhwaj2025gen2act,liang2025video,li2026turningvideomodelsgeneralist,wang2026stwamsemantictemporalworldaction}. However, these methods require full video denoising, introducing substantial inference overhead. This has motivated a more efficient line of work based on implicit future prediction, which bypasses full video denoising and directly extracts one-step features from video backbones~\citep{hu2025vpp, ma2026dit4dit,yan2026svam,pai2026mimicvideo}. One-step video features are obtained at the start of denoising process and thus retain redundant visual cues for reconstruction, making them noisy and highly entangled. Existing video-action interfaces allow the action policy to attend to these redundant visual details without explicit information constraints: ~\citet{ma2026dit4dit,pai2026mimicvideo}  directly connect one-step features to action tokens through cross-attention. \citet{hu2025vpp,yan2026svam} transform them with learned projection modules. To address this limitation, we introduce an information-constrained interface that explicitly limits the information transferred from video to action.
\vspace{-1em}
\subsection{Information Bottleneck}
\vspace{-0.5em}
The Information Bottleneck (IB) principle provides a framework for learning compact representations that preserve task-relevant information while discarding irrelevant variations \citep{tishby1999information}. Its extensions to deep learning, particularly the Variational Information Bottleneck (VIB) \citep{alemi2017deep}, make such information constraints tractable and have been widely adopted to improve representation compactness, invariance, and generalization. In robot manipulation, recent works further apply IB-style objectives to suppress redundant information in visuomotor representations and retain features relevant to action prediction \citep{bai2025rethinking,zhang2026information,pmlr-v162-du22b}. Building upon this theoretical foundation, we address a more challenging and specialized problem: learning compact and manipulation-relevant representations from highly noisy, redundant, and entangled video features. Such representations jointly encode rich visual details and dynamics, making it difficult to isolate information useful for control. This problem has remained largely unexplored in prior IB-based methods for robot manipulation.

\section{Methodology}
\label{sec:method}
As illustrated in \Figref{fig:model_arch}, we introduce IronMan, an information-constrained video-action learning framework that transforms one-step video features into compact world tokens for action generation. We first introduce the theoretical preliminaries in \Secref{sec:preliminaries}. We then present the IronMan architecture in \Secref{sec:architecture}. Finally, \Secref{sec:training} describes the joint training objective.

\subsection{Preliminaries}
\label{sec:preliminaries}
\paragraph{Problem Formulation.}
Formally, given the current observation $\vo_t$ and language instruction $\ell$, the video backbone $p_{\vtheta_{\mathrm{v}}}$ parameterized by $\vtheta_{\mathrm{v}}$ generates future frames $\vo_{t+1:t+T_v}$ and video features $\vh_{t:t+T_v}$. The video-action interface defines a latent distribution $q_{\vtheta_{\mathrm{i}}}$ parameterized by $\vtheta_{\mathrm{i}}$. During training, a sample from this distribution conditions the action policy $\pi_{\vtheta_{\mathrm{a}}}$, parameterized by $\vtheta_{\mathrm{a}}$:
\begin{equation}
\begin{aligned}
&\vo_{t+1:t+T_v},\,\vh_{t:t+T_v}
\sim
p_{\vtheta_{\mathrm{v}}}
\bigl(
\cdot \mid \vo_t,\ell
\bigr),\\
&\vz_t
\sim
q_{\vtheta_{\mathrm{i}}}
\bigl(\cdot\mid\vh_{t:t+T_v}\bigr),\\
&\rva_{t:t+T_a-1}
\sim
\pi_{\vtheta_{\mathrm{a}}}
\bigl(
\cdot \mid \ell,\vz_t
\bigr).
\end{aligned}
\label{eq:va_learning}
\end{equation}
Here, $\vz_t$ denotes the latent features used to condition the action policy, and $T_v$ and $T_a$ denote the video and action chunk lengths, respectively. Accordingly, the overall training task jointly models visual dynamics and robot actions as
\begin{equation}
\vo_{t+1:t+T_v},\,
\rva_{t:t+T_a-1}
\sim
p_{\Theta}
\bigl(
 \cdot  \mid \vo_t,\ell
\bigr),
\end{equation}
where
$\Theta=\{\vtheta_{\mathrm{v}},\vtheta_{\mathrm{i}},\vtheta_{\mathrm{a}}\}$
collects all jointly optimized model parameters, and $p_{\Theta}$ denotes the resulting joint video action model.
\vspace{-1em}
\paragraph{Flow Matching.}
We cast both video and action prediction as flow-matching problems~\citep{lipman2023flowmatchinggenerativemodeling}. Given a data sample $\vx_0\sim p_0$ and independent Gaussian noise $\epsilon\sim\mathcal{N}(\vzero,\mI)$, we define
\begin{equation}
\vx_\tau=(1-\tau)\vx_0+\tau\epsilon,
\qquad \tau\in[0,1].
\label{eq:fm_path}
\end{equation}
We train a neural vector field to predict the velocity $\epsilon-\vx_0$:
\begin{equation}
\mathcal{L}_{\mathrm{FM}}(\vtheta)
=
\E_{\vx_0,\epsilon,\tau}
\left[
\left\|
\vv_\vtheta(\vx_\tau,\tau)-(\epsilon-\vx_0)
\right\|_2^2
\right].
\label{eq:cfm_objective}
\end{equation}
At inference, we integrate the learned vector field backward from Gaussian noise at $\tau=1$ to $0$.
\vspace{-1em}

\paragraph{Information Bottleneck.}
The Information Bottleneck (IB) principle~\citep{tishby1999information}
seeks a compact representation $\mathcal{Z}$ of an input $\mathcal{X}$
that preserves information relevant to a target $\mathcal{Y}$ while
discarding irrelevant information. Formally, the IB objective is
\begin{equation}
\label{eq:ib_objective}
\min_{\mathcal{Z}} \mathrm{IB}
=
\beta\underbrace{I(\mathcal{X};\mathcal{Z})}_{\text{compression term}}
-
\underbrace{I(\mathcal{Z};\mathcal{Y})}_{\text{prediction term}}.
\end{equation}
where $I(\cdot;\cdot)$ denotes mutual information and $\beta>0$ balances compression and prediction. The compression term limits how much input information $\mathcal{Z}$ retains, while the prediction term encourages $\mathcal{Z}$ to preserve information useful for predicting $\mathcal{Y}$. Together, they favor a compact, target-relevant representation. We apply this principle to video-action learning in~\Secref{sec:training}.

\subsection{IronMan Architecture}
\label{sec:architecture}

\begin{figure}[htbp]
    \centering
    \includegraphics[width=\linewidth]{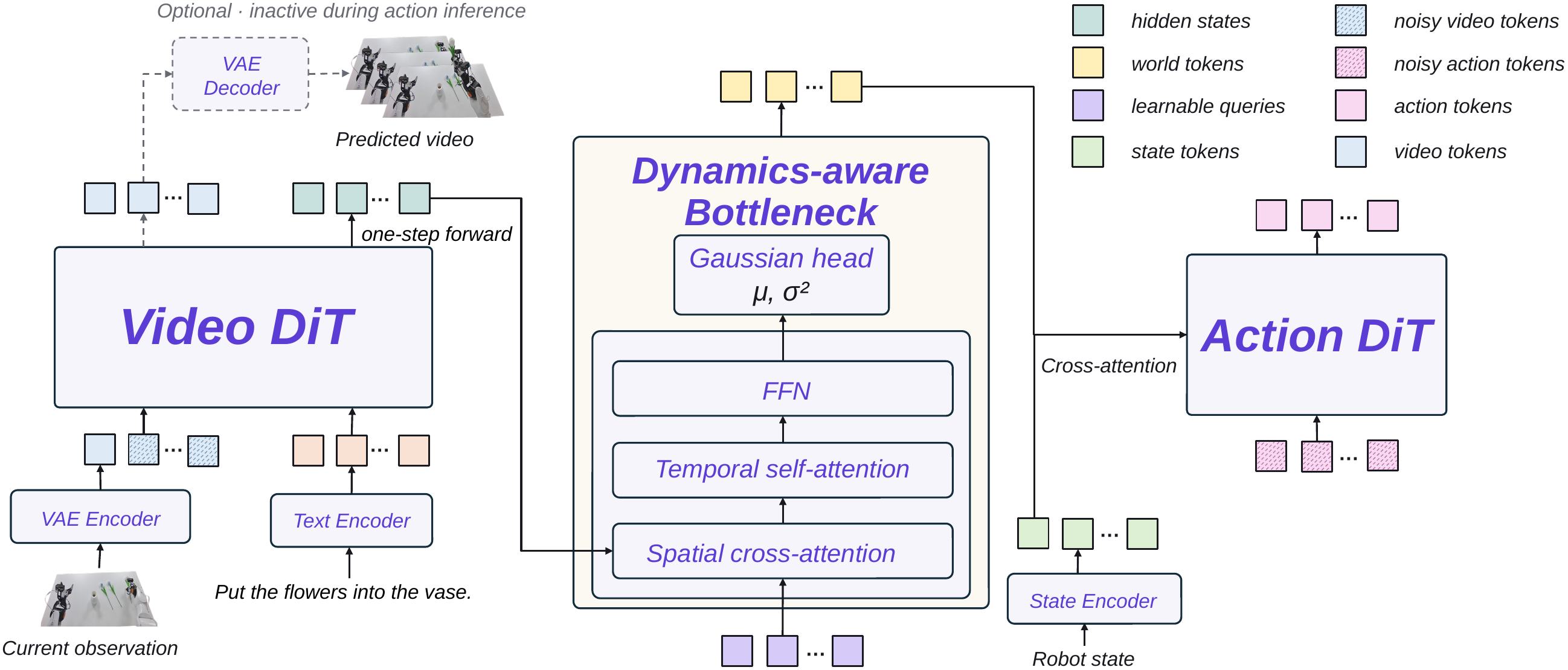}
    \caption{Overview of IronMan. A dynamics-aware bottleneck transforms one-step video features into compact world tokens using a spatio-temporal transformer and a Gaussian head. The action policy attends to world tokens and robot state tokens to generate actions; video decoding is optional and inactive during action inference.}
    \label{fig:model_arch}
\end{figure}

As shown in \Figref{fig:model_arch}, IronMan comprises three key components: a video diffusion transformer~(Video DiT) that models visual dynamics; a dynamics-aware bottleneck that compresses the resulting features into compact world tokens under an information constraint; and an action diffusion transformer~(Action DiT) that generates robot actions conditioned on these world tokens.

\paragraph{Video DiT.}
We use Wan2.2-5B~\citep{wan2025} as our video backbone $p_{\vtheta_{\mathrm{v}}}$. The VAE encodes the current observation $\vo_t$ into
conditioning latents, while a text encoder embeds the language
instruction $\ell$. Conditioned on these inputs, the Video DiT processes
noisy future video latents and predicts the velocity field introduced
in \Secref{sec:preliminaries}. To obtain video features, we extract the hidden states
$\vh_{t:t+T_v}$ from layer $L_{\mathrm{feat}}$ in a one-step forward process. 
\vspace{-1em}
\paragraph{Dynamics-aware Bottleneck.}
The dynamics-aware bottleneck extracts action-relevant dynamics from highly noisy video features then compresses them into compact world tokens. It consists of a spatio-temporal aggregation module and a Gaussian head that parameterizes $q_{\vtheta_{\mathrm{i}}}(\vz_t\mid\vh_{t:t+T_v})$.

The aggregation module uses learnable queries $\mQ$ to summarize video features through stacked Qformer-style blocks~\citep{li2023blip}. Each block performs spatial cross-attention to video tokens, temporal self-attention across query tokens, and a feed-forward update:
\begin{equation}
\mQ^{(l+1)}
=
\operatorname{FFN}\!\left(
\operatorname{SelfAttn}_{\mathrm{temp}}\!\left(
\operatorname{CrossAttn}_{\mathrm{spatial}}\!
\left(\mQ^{(l)},\vh_{t:t+T_v}\right)
\right)
\right),
\label{eq:spatiotemporal_aggregation}
\end{equation}

where $\mQ^{(0)}=\mQ$. A shared Gaussian head maps the final query tokens $\mQ^{(L)}$, where $L$ is the number of aggregation blocks, to a mean $\vmu_t$ and a diagonal covariance $\mSigma_t$, defining
\begin{equation}
q_{\vtheta_{\mathrm{i}}}(\vz_t\mid\vh_{t:t+T_v})
=
\mathcal{N}(\vz_t;\vmu_t,\mSigma_t).
\label{eq:world_token_distribution}
\end{equation}
During training, world tokens are sampled via reparameterization,
$\vz_t=\vmu_t+\mSigma_t^{1/2}\rvepsilon$,
where $\rvepsilon\sim\mathcal{N}(\vzero,\mI)$; at inference, we use $\vz_t=\vmu_t$.
\vspace{-1em}
\paragraph{Action DiT.}
We instantiate the action policy $\pi_{\vtheta_{\mathrm{a}}}$ as a separate diffusion transformer that generates an action chunk of length $T_a$. Conditioned on the action flow timestep $\tau_a$, it applies self-attention to noisy action tokens and cross-attention to world tokens $\vz_t$ and robot state embeddings. The instruction $\ell$ is incorporated through the instruction-conditioned video features encoded in $\vz_t$. The model predicts an action velocity field introduced
in \Secref{sec:preliminaries}, which is integrated from Gaussian noise at $\tau_a=1$ to $0$ to generate $\rva_{t:t+T_a-1}$.

\subsection{Training Objective}
\label{sec:training}

\paragraph{Action Learning with Variational Information Bottleneck.}
We instantiate the information bottleneck objective in \eqref{eq:ib_objective} with video features $\vh_{t:t+T_v}$ as the input, world tokens $\vz_t$ as the compressed latents, and the action chunk $\rva_{t:t+T_a-1}$ as the prediction target. The resulting objective balances retaining action-relevant information against compressing video features. Using a standard Gaussian prior $p(\vz_t)=\mathcal{N}(\vzero,\mI)$, the compression term admits the variational upper bound~\citep{alemi2017deep}
\begin{equation}
I(\vh_{t:t+T_v};\vz_t)
\leq
\mathbb{E}_{\vh_{t:t+T_v}}
\left[
D_{\mathrm{KL}}\!\left(
q_{\vtheta_{\mathrm{i}}}(\vz_t\mid\vh_{t:t+T_v})
\,\|\,p(\vz_t)
\right)
\right].
\label{eq:ib_compression_bound}
\end{equation}
Maximizing the prediction term $I(\vz_t;\rva_{t:t+T_a-1})$ corresponds to minimizing conditional action uncertainty. We use the action flow-matching loss as a tractable surrogate for this prediction term, yielding the action-centric objective
\begin{equation}
\mathcal{L}_{\mathrm{ac}}(\vtheta_\mathrm{i}, \vtheta_\mathrm{a})
=
\mathcal{L}_{\mathrm{FM}}(\vtheta_{\mathrm{a}})
+
\beta\,
\mathbb{E}_{\vh_{t:t+T_v}}
\left[
D_{\mathrm{KL}}\!\left(
q_{\vtheta_{\mathrm{i}}}(\vz_t\mid\vh_{t:t+T_v})
\,\|\,p(\vz_t)
\right)
\right].
\label{eq:action_centered_objective}
\end{equation}
The action-centric loss captures the fundamental trade-off between action prediction and compressing visual-dynamics information.
\vspace{-1em}
\paragraph{Information-Constrained Video-Action Learning.}
To achieve simultaneous modeling of
latent representations for both video and action, we complement the action-centered objective with dense visual supervision. We jointly optimize $\vtheta_{\mathrm{v}}$, $\vtheta_{\mathrm{i}}$, and $\vtheta_{\mathrm{a}}$ using
\begin{equation}
\mathcal{L}
=
\mathcal{L}_{\mathrm{FM}}(\vtheta_{\mathrm{v}}) + 
\mathcal{L}_{\mathrm{ac}}(\vtheta_\mathrm{i}, \vtheta_\mathrm{a})
.
\label{eq:total_training_objective}
\end{equation}
This joint objective trains the video backbone to model rich visual dynamics, while the information bottleneck balances action prediction against visual information compression, encouraging world tokens to preserve control-relevant dynamics and discard details unnecessary for action prediction.
The training and inference procedures are detailed in \Appref{app:algorithms}.

\section{Simulation Experiments}
We evaluate IronMan on RoboTwin Clean dataset~\citep{chen2025robotwin} and the standard LIBERO dataset~\citep{liu2023libero}, each with an in-domain (ID) benchmark and a corresponding out-of-distribution (OOD) benchmark, to assess its ID performance and OOD robustness. We further conduct ablations to validate individual modules, empirically analyze the bottleneck mechanism in IronMan, and provide visualizations to illustrate the effectiveness of this paradigm.

\subsection{Setup}
\paragraph{Datasets.} We use two simulation datasets: RoboTwin Clean and LIBERO. RoboTwin Clean comprises 50 manipulation tasks from RoboTwin 2.0, with 50 demonstrations per task collected under the \texttt{demo\_clean} setting. Standard LIBERO datasetcomprises 40 manipulation tasks across four suites: LIBERO-Spatial, LIBERO-Object, LIBERO-Goal, and LIBERO-Long. Training follows a multi-task setting, with a single policy trained per dataset.
\vspace{-1em}
\paragraph{Benchmarks.} We evaluate IronMan on four benchmarks.
For RoboTwin, clean2clean (c2c) evaluates policies trained on clean demonstrations under the same \texttt{demo\_clean} setting as training, while clean2random (c2r) evaluates them under the \texttt{demo\_randomized} setting, with 100 evaluation episodes per task across all 50 tasks.
For LIBERO, we use the standard benchmark for ID evaluation, with 50 evaluation episodes per task across all 40 tasks, and LIBERO-Plus~\citep{fei2026liberoplus} for OOD evaluation, with 10,030 evaluation episodes in total. We report the average success rate for each benchmark.
\vspace{-1em}
\paragraph{Baselines.} We compare IronMan with two categories of baselines:
Vision-Language-Action Models (VLAs), including
$\pi_{0.5}$~\citep{physicalintelligence2025pi05}, X-VLA~\citep{zheng2025xvlasoftpromptedtransformerscalable}, Abot-M0\citep{yang2026abotm0vlafoundationmodel}, and
StarVLA~\citep{community2026starvla};
and Video Action Models (VAMs), including
FastWAM, FastWAM-Joint, FastWAM-IDM~\citep{yuan2026fastwam},
Motus~\citep{bi2026motus}, LingBot-VA~\citep{li2026lingbotva}, OpenWAM-Joint~\citep{wang2026openwam}, and DiT4DiT~\citep{ma2026dit4dit}.
\vspace{-1em}
\paragraph{Implementation Details.}
We initialize the Video DiT from Wan2.2-5B~\citep{wan2025}, which is then finetuned on the two simulation datasets.
Our dynamics-aware bottleneck uses a three-layer spatio-temporal transformer with a hidden dimension of 512.
Following the architecture design of~\citet{yuan2026fastwam}, our Action DiT contains approximately 1B parameters, with a hidden dimension of 1024. We set the action horizon \(T_a\) to 32 and temporally downsample the video frames by a factor of 4. The latency of our model is measured on a NVIDIA RTX 5090~(24G) GPU.
More implementation details are provided in~\Appref{app:training_details}.

\subsection{In-Domain Evaluation Results}

We evaluate IronMan’s in-domain performance on LIBERO and RoboTwin clean2clean (c2c) in Table~\ref{tab:id_results}. Without embodied pretraining, IronMan achieves the highest  success rate on LIBERO. Notably, IronMan achieves 99.0\% on LIBERO-Long, highlighting its effectiveness on long-horizon manipulation tasks. On RoboTwin c2c, IronMan achieves 79.4\%, exceeding the strongest reported baseline, FastWAM. These results demonstrate IronMan’s strong in-domain performances.

\begin{table}[t]
\centering
\caption{In-domain performance on LIBERO and RoboTwin
clean2clean (c2c). Success rates are reported in percentages.
Em.\ PT.\ denotes embodied pretraining.
Bold and underlined values indicate the best and second-best
results in each column, respectively.}
\label{tab:id_results}

\small
\setlength{\tabcolsep}{4pt}
\renewcommand{\arraystretch}{1.08}

\begin{tabular}{
l | >{\centering\arraybackslash}p{2.4em} | ccccc | c
}
\toprule
\multirow{2}{*}{Models}
& \multirow{2}{*}{\shortstack{Em.\\PT.}}
& \multicolumn{5}{c|}{LIBERO}
& \multirow{2}{*}{\shortstack{RoboTwin\\c2c}} \\
& & Spatial & Object & Goal & Long & Avg. & \\
\midrule

$\pi_{0.5}$
& \cmark
& \underline{98.8} & 98.2 & 98.0
& 92.4 & 96.9 & 70.7 \\

X-VLA
& \cmark
& 98.2 & 98.6 & 97.8
& 97.6 & 98.1 & 68.0 \\

Abot-M0
& \cmark
& \underline{98.8} & \underline{99.8} & \textbf{99.0}
& 96.6 & \underline{98.6} & 57.4 \\

starVLA
& \xmark
& 98.7 & 99.7 & \underline{98.6}
& 94.2 & 97.8 & 46.5 \\

FastWAM
& \xmark
& 98.2 & \textbf{100.0} & 97.0
& 95.2 & 97.6 & \underline{77.8} \\

LingBot-VA
& \cmark
& 98.5 & 99.6 & 97.2
& \underline{98.5} & 98.5 & -- \\

Motus
& \cmark
& 96.8 & \underline{99.8} & 96.6
& 97.6 & 97.7 & -- \\

\rowcolor{blue!8}
\textbf{IronMan}
& \xmark
& \textbf{99.0} & \textbf{100.0} & 97.8
& \textbf{99.0} & \textbf{99.0} & \textbf{79.4} \\
\bottomrule
\end{tabular}
\end{table}

\subsection{Out-of-Distribution Evaluation Results}

For a fair comparison, we compare IronMan with VAM baselines
without large-scale embodied pretraining and report
inference latency without engineering optimizations.
As shown in~\Tabref{tab:libero_plus_robotwin_c2r},
IronMan achieves an overall success rate of 79.1\% on LIBERO-Plus,
surpassing the strongest baseline, DiT4DiT, by 10.4\%.
It ranks first across all seven perturbation categories,
with the largest gains over the best competing results
in sensor noise (+18.3\%) and background
textures (+10.2\%).
On RoboTwin c2r, IronMan ranks second with a success rate of 12.1\%,
trailing OpenWAM-Joint by 1.1\% with less than half
its inference latency.
 These results demonstrate IronMan's strong generalization under visual and physical shifts while maintaining inference efficiency.

\begin{table*}[ht]
\centering
\caption{Performance on LIBERO-Plus and RoboTwin clean2random (c2r),
and inference latency.
Success rates are reported in percentages.
Bold and underlined success rates indicate the best and second-best
results in each column, respectively.}
\label{tab:libero_plus_robotwin_c2r}
\footnotesize
\setlength{\tabcolsep}{1.5pt}
\renewcommand{\arraystretch}{1.15}
\begin{tabular}{l | *{8}{c} | c | c}
\toprule
\multirow{2}{*}{Models}
& \multicolumn{8}{c|}{LIBERO-Plus}
& \multirow{2}{*}{\shortstack{RoboTwin\\c2r}}
& \multirow{2}{*}{\shortstack{Latency\\(ms)}} \\
& Camera & Robot & Language & Light & Background & Noise & Layout & Overall & & \\
\midrule

DiT4DiT
& \underline{59.7} & 61.1 & 88.4 & 90.8 & 33.7 & \underline{60.0}
& \underline{83.6} & \underline{68.7} & -- & 180 \\

FastWAM
& 16.8 & 43.7 & 68.1 & 77.2 & 51.5 & 36.5 & 59.5 & 49.2 & 1.9 & 190 \\
FastWAM-Joint
& 41.3 & 65.5 & \underline{91.0} & 86.4 & 53.8 & 56.2 & 79.2 & 67.4 & 5.2 & 580 \\
FastWAM-IDM
& 37.7 & \underline{67.2} & 90.3 & \underline{92.7}
& \underline{54.2} & 56.6 & 79.2 & 67.9 & 9.3 & 810 \\

OpenWAM-Joint
& 1.2 & 53.4 & 71.1 & 77.3 & 43.8 & 18.4 & 52.2 & 43.7
& \textbf{13.2} & 429 \\
\rowcolor{blue!8}
\textbf{IronMan}
& \textbf{64.4} & \textbf{75.9} & \textbf{92.7} & \textbf{93.4}
& \textbf{64.4} & \textbf{78.3} & \textbf{83.7} & \textbf{79.1}
& \underline{12.1} & 200 \\
\bottomrule
\end{tabular}
\end{table*}

\subsection{Ablation Study}
\Tabref{tab:module_ablation} disentangles the contributions of the dynamics-aware bottleneck (DB) architecture and the information bottleneck (IB) objective. The baseline directly conditions action generation on hidden states extracted from the fixed feature layer $L_{\mathrm{feat}}$. The IB-only variant applies a shared Gaussian head to each video token, with a wider MLP at the same depth to ensure comparable model capacity. The DB-only variant sets $\beta=0$, removing the information constraint. Applying IB directly to raw video features yields no improvement over the baseline (73.2\% vs.\ 73.4\%), suggesting that token-wise compression alone is insufficient to improve OOD performance. DB alone achieves a modest gain to 74.3\%, while combining DB with IB raises the overall score to 79.1\%. These results support the complementary roles of structured dynamics extraction and explicit information constraints.

\begin{table*}[ht]
\centering
\caption{Ablation study of the dynamics-aware bottleneck (DB) architecture and information bottleneck
(IB) objective across seven types of perturbations in LIBERO-Plus.
Bold values indicate the best result in each column.}
\label{tab:module_ablation}

\footnotesize
\setlength{\tabcolsep}{2pt}
\renewcommand{\arraystretch}{1.15}

\begin{tabular}{@{} cc *{8}{c} @{}}
\toprule
\shortstack{DB arch.}
& \shortstack{IB obj.}
& \shortstack{Camera}
& \shortstack{Robot}
& \shortstack{Language}
& \shortstack{Light}
& \shortstack{Background}
& \shortstack{Noise}
& \shortstack{Layout}
& Overall \\
\midrule
\xmark & \xmark
& 49.8 & \textbf{77.1} & 93.2 & 87.0
& 63.3 & 61.3 & 82.9 & 73.4 \\

\xmark & \cmark
& 56.0 & 73.6 & 90.8 & 83.1
& \textbf{65.5} & 61.7 & 82.0 & 73.2 \\

\cmark & \xmark
& 54.7 & 75.9 & \textbf{93.4} & 85.2
& 63.9 & 64.2 & 82.8 & 74.3 \\

\rowcolor{blue!8}
\cmark & \cmark
& \textbf{64.4} & 75.9 & 92.7 & \textbf{93.4}
& 64.4 & \textbf{78.3} & \textbf{83.7}
& \textbf{79.1} \\
\bottomrule
\end{tabular}
\end{table*}

\subsection{Bottleneck Mechanism Analysis}

\begin{figure}[htbp]
    \centering
    \begin{subfigure}[t]{0.4\linewidth}
        \centering
        \includegraphics[width=\linewidth]{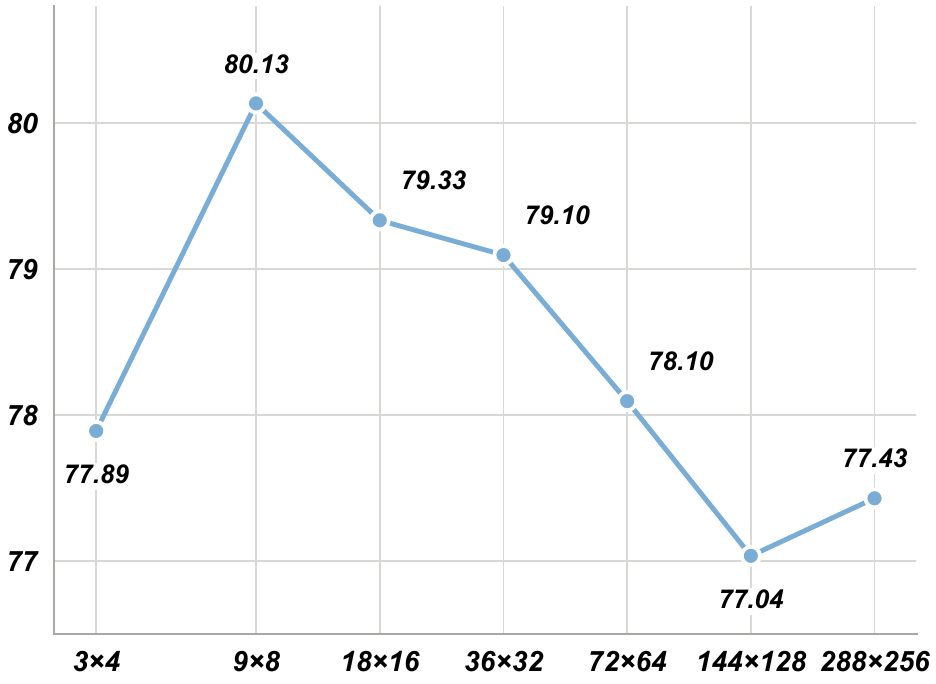}
        \caption{Bottleneck size $Q \times D$.}
        \label{fig:bottleneck_capacity_libero_plus}
    \end{subfigure}\hspace{0.04\linewidth}
    \begin{subfigure}[t]{0.4\linewidth}
        \centering
        \includegraphics[width=\linewidth]{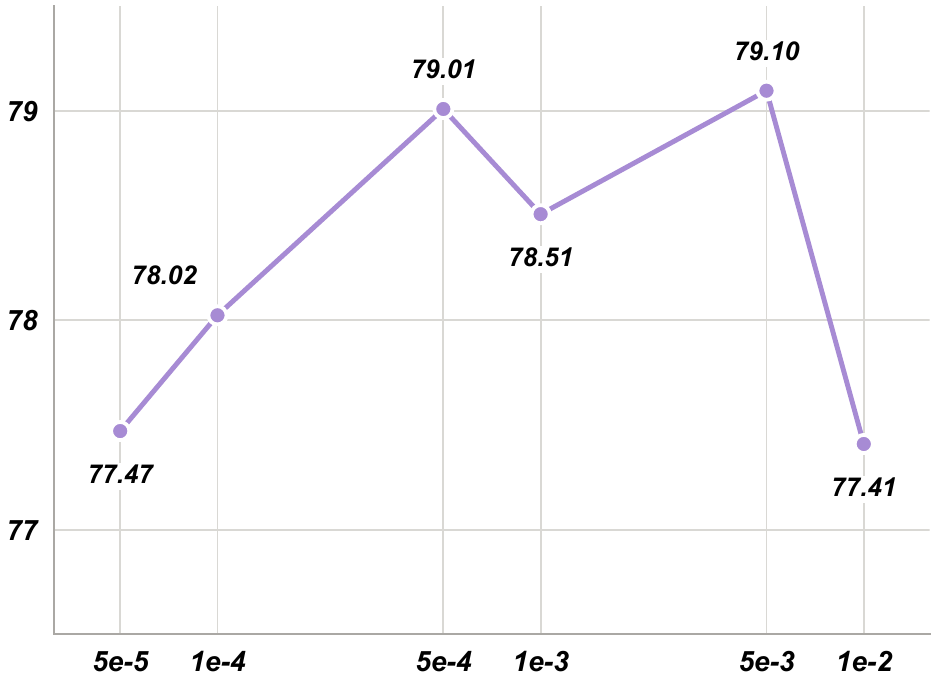}
        \caption{Constraint strength $\beta$.}
        \label{fig:ib_beta_sweep_libero_plus}
    \end{subfigure}
    \caption{Bottleneck mechanism analysis on LIBERO-Plus. Success rates (\%) are shown for different bottleneck capacities $Q \times D$ and information bottleneck constraint strengths $\beta$.}
    \label{fig:bottleneck_mechanism_analysis}
\end{figure}

We examine how bottleneck size ($Q \times D$) and constraint strength ($\beta$) affect IronMan’s OOD performance, where $Q$ and $D$ denote the number and dimensionality of world tokens, respectively. All configurations outperform the 73.4\% baseline in~\Tabref{tab:module_ablation}. \Figref{fig:bottleneck_capacity_libero_plus} shows that smaller bottleneck limits the model’s representation capacity, whereas larger bottlenecks may expose the action head to excessive irrelevant visual details. Similarly, \Figref{fig:ib_beta_sweep_libero_plus} shows that intermediate $\beta$ values perform best, suggesting that weak constraints allow irrelevant information to pass through, while overly strong constraints discard information needed for control. Together, these results suggest a trade-off between retaining sufficient information for action prediction and compressing irrelevant visual details, consistent with the principle of bottleneck mechanism.

\subsection{Visualization Analysis}

\begin{figure}[htbp]
    \centering
    \includegraphics[width=\linewidth]{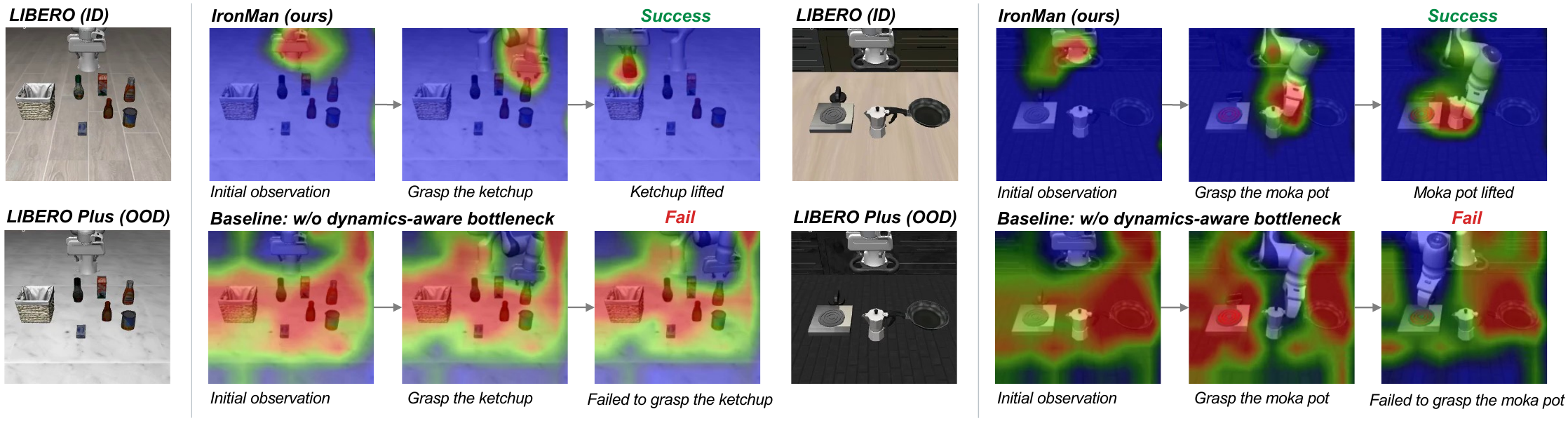}
    \caption{Comparison of attention heatmaps for IronMan and the baseline without the dynamics-aware bottleneck under a background shift from LIBERO (ID) to LIBERO Plus (OOD).}
    \label{fig:sim_vis_main}
\end{figure}

\Figref{fig:sim_vis_main} compares attention heatmaps of IronMan and the baseline without the dynamics-aware bottleneck under a background shift from LIBERO to LIBERO Plus. Across the illustrated reaching and grasping stages, IronMan focuses more selectively on the gripper–object interaction region, whereas the baseline also attends broadly to task-irrelevant backgrounds. This pattern is consistent with the goal of information-constrained video-action learning: preserving control-relevant information while limiting reliance on irrelevant visual details. Additional visualizations are provided in~\Appref{sec:attn_vis}, and intervention experiments further assess robustness to visual perturbations in~\Appref{app:intervention}.

\section{Real-World Experiments}
\begin{figure}[htbp]
    \centering
    \includegraphics[width=\linewidth]{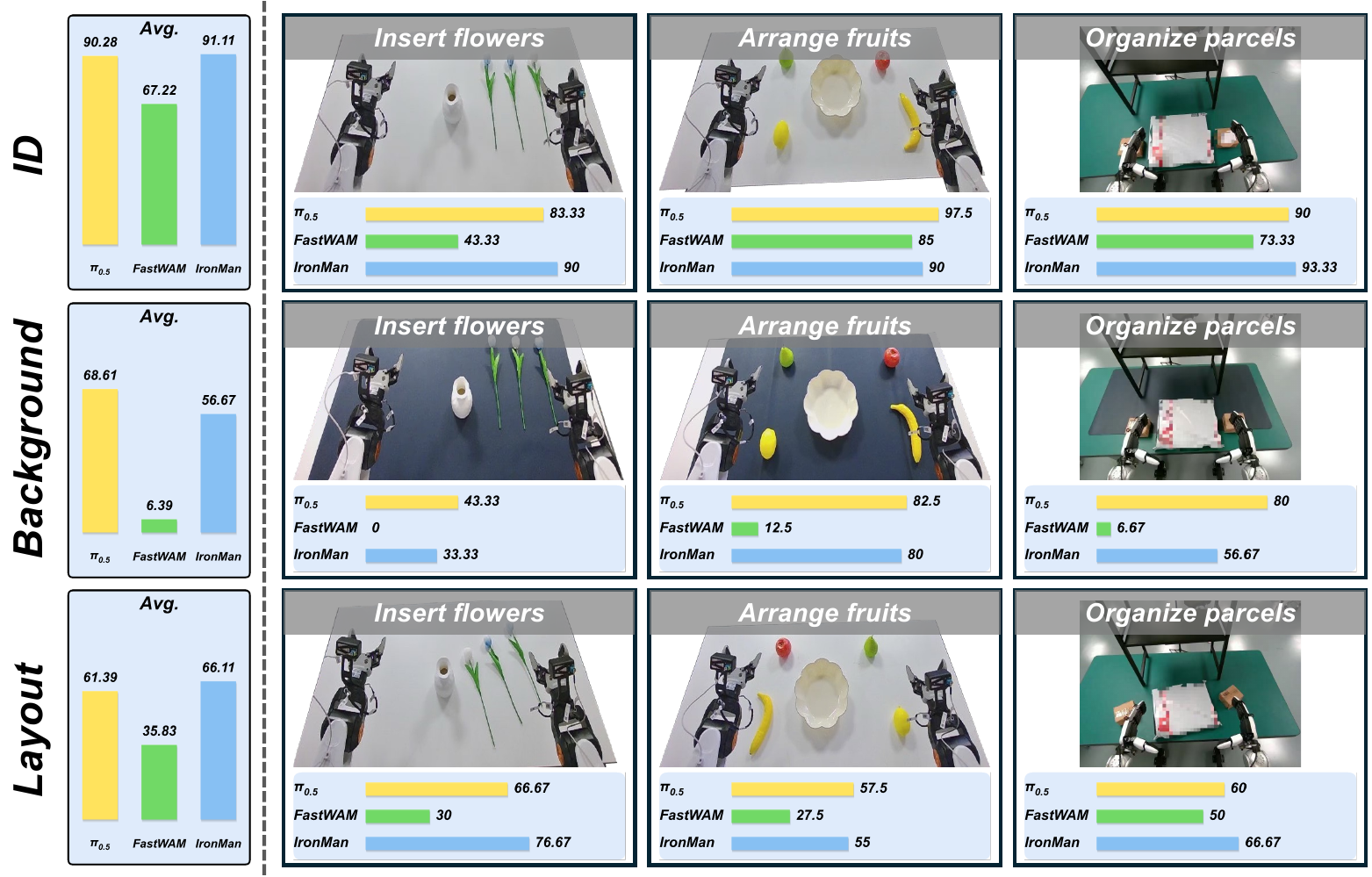}
    \caption{Real-world performance under ID, Background, and Layout settings.}
    \label{fig:real_world_exp}
\end{figure}

\paragraph{Setup.} To evaluate IronMan’s capabilities on real-world tasks, we design three tasks across the AGIBOT G2  and ROBOTERA M7 platforms: \textbf{insert flowers}, \textbf{arrange fruits}, and \textbf{organize parcels}.  These tasks evaluate the model’s capabilities in spatial perception, fine-grained manipulation, and long-horizon task execution. For each task, we collect 100 demonstrations via teleoperation. Task demonstrations can be found in~\Appref{sec:real_task_demo}.
\vspace{-1em}
\paragraph{Baselines.}
We select  $\pi_{0.5}$~\citep{physicalintelligence2025pi05} and 
FastWAM~\citep{yuan2026fastwam} as our baselines. All baselines are trained using their official training codes.
\vspace{-1em}
\paragraph{Evaluation Protocol.} For each task, we design three evaluation settings: \textbf{ID}, where the scene configuration follows the same distribution as the training data; \textbf{Background}, where we alter the tabletop texture to introduce strong visual distractions; and \textbf{Layout}, where we modify the positions and orientations of objects. For each task under each setting, we conduct 10 trials. Each trial receives a score from 0 to 10 according to the score protocol in~\Appref{app:score}.
\vspace{-1em}
\paragraph{Results.}
\Figref{fig:real_world_exp} summarizes the real-world results. Under ID setting, IronMan achieves the highest average normalized score of 91.11, comparable to $\pi_{0.5}$ at 90.28 and exceeding FastWAM by 23.89. Background perturbations substantially degrade FastWAM’s performance, particularly on flower insertion task requiring precise manipulation. IronMan achieves an average score of 56.67 and outperforms FastWAM by 50.28, supporting greater robustness under background shift. Under layout shifts, IronMan achieves the highest average score of 66.11, surpassing $\pi_{0.5}$ and FastWAM by 4.72 and 30.28, respectively. These results show that, even without the $\pi_{0.5}$-scale embodied pretraining, IronMan achieves comparable real-world robustness and substantially outperforms FastWAM.

\begin{figure}[ht]
    \centering
    \includegraphics[width=\linewidth]{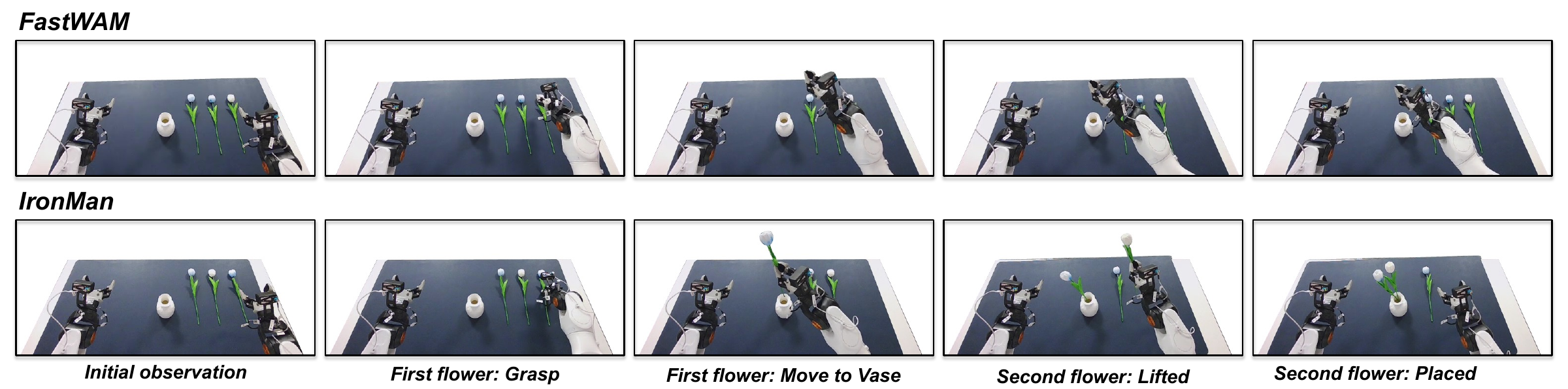}
    \caption{Comparison of the flower insertion task under background shift.}
    \label{fig:real_world_case}
\end{figure}

\vspace{-1em}
\paragraph{Case Analysis.}
As shown in~\Figref{fig:real_world_case}, IronMan  accurately grasps the target flowers and transfers them to the vase. In contrast, FastWAM fails to establish a secure grasp. The result indicates our model maintains more accurate grasping and placement than FastWAM under this background shift in this example.

\section{Conclusions}

We propose IronMan, an information-constrained video-action learning framework that aligns visual representations with robot control. Its dynamics-aware bottleneck combines spatio-temporal aggregation and information regularization to distill one-step video features into compact world tokens. IronMan achieves strong in-domain performance and also demonstrates strong OOD robustness while maintaining efficient inference. Ablations validate the complementary roles of aggregation and regularization, and real-world evaluations further support its robustness.

\bibliography{iclr2027_conference}
\bibliographystyle{iclr2027_conference}

\newpage
\appendix
\section{Appendix}
\subsection{Model and Training Details}
\label{app:details}

\subsubsection{Training and inference algorithms}
\label{app:algorithms}

\Algref{alg:ironman_training} summarizes the training process, and algorithm~\ref{alg:ironman_inference}
describes action inference. We use $\vx_0$ for clean video latents,
$\rva_{t:t+T_a-1}$ for the target action chunk, $\vc_t$ for
current-observation latents, and $\vs_t$ for robot state embeddings.
$\mathrm{VAE}$, $\mathrm{Enc}_{\mathrm{text}}$, and $\mathrm{Enc}_{\mathrm{state}}$ denote
the corresponding encoders. The video and action velocity predictors are
$\vv_{\vtheta_{\mathrm{v}}}$ and $\vv_{\vtheta_{\mathrm{a}}}$.
The dynamics-aware bottleneck combines the spatio-temporal aggregation in
\eqref{eq:spatiotemporal_aggregation} with the Gaussian head in
\eqref{eq:world_token_distribution}.

During training, video flow matching uses noisy ground-truth video latents,
while a separate one-step forward pass extracts action-conditioning features
from Gaussian future latents and the current observation. Both passes share
the Video DiT parameters. The latter pass is identical during action inference;
world tokens are then computed once and reused throughout action denoising.
The conditioning frame
is excluded from the video loss. The implementation averages the Gaussian KL
over world tokens.

\begin{algorithm}[htbp]
\caption{IronMan training}
\label{alg:ironman_training}
\begin{algorithmic}[1]
\Require Dataset $\mathcal{D}$; parameters $\Theta=\{\vtheta_{\mathrm{v}},\vtheta_{\mathrm{i}},\vtheta_{\mathrm{a}}\}$; feature layer $L_{\mathrm{feat}}$; IB weight $\beta$
\While{not converged}
    \State Sample a minibatch $(\vo_{t:t+T_v},\ell,\text{robot state},\rva_{t:t+T_a-1})$ from $\mathcal{D}$
    \State $\vx_0\gets \mathrm{VAE}(\vo_{t:t+T_v})$; $\vc_t\gets$ first temporal slice of $\vx_0$
    \State $\ve_\ell\gets \mathrm{Enc}_{\mathrm{text}}(\ell)$; $\vs_t\gets \mathrm{Enc}_{\mathrm{state}}(\text{robot state})$
    \State Sample video training time $\tau$ and $\rvepsilon_{\mathrm{v}}\sim\mathcal{N}(\vzero,\mI)$
    \State $\vx_\tau\gets(1-\tau)\vx_0+\tau\rvepsilon_{\mathrm{v}}$; replace its first temporal slice with $\vc_t$
    \State $\hat{\vv}_{\mathrm{v}}\gets\vv_{\vtheta_{\mathrm{v}}}(\vx_\tau,\tau\mid\ve_\ell)$
    \State $\mathcal{L}_{\mathrm{FM}}(\vtheta_{\mathrm{v}})\gets\|\hat{\vv}_{\mathrm{v}}-(\rvepsilon_{\mathrm{v}}-\vx_0)\|_2^2$ on future latents
    \State Sample Gaussian video latents; replace their first temporal slice with $\vc_t$
    \State $\vh_{t:t+T_v}\gets$ extract hidden states from $L_{\mathrm{feat}}$ in a one-step  forward on these latents, 
    \State $(\vmu_t,\mSigma_t)\gets\operatorname{DynamicsAwareBottleneck}_{\vtheta_{\mathrm{i}}}(\vh_{t:t+T_v})$
    \State Sample $\rvepsilon_{\mathrm{z}}\sim\mathcal{N}(\vzero,\mI)$; $\vz_t\gets\vmu_t+\mSigma_t^{1/2}\rvepsilon_{\mathrm{z}}$
    \State $\mathcal{L}_{\mathrm{KL}}\gets\KL\!\left(\mathcal{N}(\vmu_t,\mSigma_t)\,\|\,\mathcal{N}(\vzero,\mI)\right)$
    \State Sample action training time $\tau_a$ and $\rvepsilon_{\mathrm{a}}\sim\mathcal{N}(\vzero,\mI)$
    \State $\va_{\tau_a}\gets(1-\tau_a)\rva_{t:t+T_a-1}+\tau_a\rvepsilon_{\mathrm{a}}$
    \State $\hat{\vv}_{\mathrm{a}}\gets\vv_{\vtheta_{\mathrm{a}}}(\va_{\tau_a},\tau_a\mid\vz_t,\vs_t)$
    \State $\mathcal{L}_{\mathrm{FM}}(\vtheta_{\mathrm{a}})\gets\|\hat{\vv}_{\mathrm{a}}-(\rvepsilon_{\mathrm{a}}-\rva_{t:t+T_a-1})\|_2^2$
    \State $\mathcal{L}\gets\mathcal{L}_{\mathrm{FM}}(\vtheta_{\mathrm{v}})+\mathcal{L}_{\mathrm{FM}}(\vtheta_{\mathrm{a}})+\beta\mathcal{L}_{\mathrm{KL}}$
    \State Update $\Theta$ by backpropagating the minibatch mean of $\mathcal{L}$ through both passes
\EndWhile
\State \Return $\Theta$
\end{algorithmic}
\end{algorithm}

\begin{algorithm}[htbp]
\caption{IronMan action inference}
\label{alg:ironman_inference}
\begin{algorithmic}[1]
\Require Observation $\vo_t$; instruction $\ell$; robot state; trained $\Theta$; feature layer $L_{\mathrm{feat}}$
\Require Horizons $T_v,T_a$; action schedule $1=\tau_a^{(0)}>\cdots>\tau_a^{(N_{\mathrm{infer}})}=0$
\State $\vc_t\gets \mathrm{VAE}(\vo_t)$; $\ve_\ell\gets \mathrm{Enc}_{\mathrm{text}}(\ell)$; $\vs_t\gets \mathrm{Enc}_{\mathrm{state}}(\text{robot state})$
\State Sample Gaussian video latents for horizon $T_v$; replace their first temporal slice with $\vc_t$
\State $\vh_{t:t+T_v}\gets$ extract hidden states from $L_{\mathrm{feat}}$ in a one-step forward on these latents
\State $(\vmu_t,\mSigma_t)\gets\operatorname{DynamicsAwareBottleneck}_{\vtheta_{\mathrm{i}}}(\vh_{t:t+T_v})$
\State $\vz_t\gets\vmu_t$
\State Sample $\rvepsilon_{\mathrm{a}}\sim\mathcal{N}(\vzero,\mI)$ for $T_a$ actions; $\hat{\va}^{(0)}\gets\rvepsilon_{\mathrm{a}}$
\For{$k=0,\ldots,N_{\mathrm{infer}}-1$}
    \State $\hat{\vv}_{\mathrm{a}}\gets\vv_{\vtheta_{\mathrm{a}}}(\hat{\va}^{(k)},\tau_a^{(k)}\mid\vz_t,\vs_t)$
    \State $\hat{\va}^{(k+1)}\gets\hat{\va}^{(k)}+(\tau_a^{(k+1)}-\tau_a^{(k)})\hat{\vv}_{\mathrm{a}}$
\EndFor
\State \Return Action chunk $\hat{\rva}_{t:t+T_a-1}\gets\hat{\va}^{(N_{\mathrm{infer}})}$
\end{algorithmic}
\end{algorithm}

\FloatBarrier

\subsubsection{Training Configurations}
\label{app:training_details}

We report the video generation training and the subsequent
video-action learning stage separately for each dataset. We first
finetune the Wan2.2-5B~\citep{wan2025} video backbone using the video flow-matching loss on the simulation dataset. Then we train the Video DiT, dynamics-aware bottleneck, and Action DiT together
using the objective in \eqref{eq:total_training_objective}.

\paragraph{LIBERO.}
We jointly train on the four LIBERO suites. The external and wrist camera
images are each resized to $224\times224$ and concatenated horizontally into
a $224\times448$ video frame.  The details are shown in~\Tabref{tab:libero_training_config}.


\paragraph{RoboTwin Clean.}
We train on the 50-task  dataset collected under \texttt{demo\_clean} setting. The head,
left-wrist, and right-wrist camera images are resized to $240\times320$
before the dataset's multi-camera composition produces a $384\times320$
video frame. 
\Tabref{tab:robotwin_training_config} gives the full configurations.

\begin{table*}[ht]
\centering
\caption{Training configurations on LIBERO and RoboTwin Clean.}
\label{tab:training_config}
\scriptsize
\setlength{\tabcolsep}{3pt}
\renewcommand{\arraystretch}{0.95}
\captionsetup[subtable]{font=small,skip=3pt}

\begin{minipage}[t]{0.49\textwidth}
\centering
\subcaption{LIBERO}
\label{tab:libero_training_config}
\begin{tabular}{@{}lcc@{}}
\toprule
Setting & video generation & video-action learning\\
\midrule
Global batch size          & 128                & 256                \\
Training steps             & 10k                & 30k                \\
Learning rate              & $10^{-4}$          & $10^{-4}$          \\
Video resolution
                           & $224\times448$     & $224\times448$     \\
Sampled video frames       & 9                  & 9                  \\
Action chunk length $T_a$  & ---                & 32                 \\
Number of queries $Q$      & ---                & 36                 \\
World token dimension $D$  & ---                & 32                 \\
IB weight $\beta$          & ---                & $5\times10^{-3}$   \\
Feature layer $L_{\mathrm{feat}}$
                           & ---                & 17                 \\
Bottleneck hidden dimension & ---               & 512                \\
\bottomrule
\end{tabular}
\end{minipage}
\hfill
\begin{minipage}[t]{0.49\textwidth}
\centering
\subcaption{RoboTwin Clean}
\label{tab:robotwin_training_config}
\begin{tabular}{@{}lcc@{}}
\toprule
Setting & video generation & video-action learning\\
\midrule
Global batch size          & 128                & 256                \\
Training steps             & 10k                & 30k                \\
Learning rate              & $10^{-4}$          & $10^{-4}$          \\
Video resolution ($H\times W$)
                           & $384\times320$     & $384\times320$     \\
Sampled video frames       & 9                  & 9                  \\
Action chunk length $T_a$  & ---                & 32                 \\
Number of queries $Q$      & ---                & 36                 \\
World token dimension $D$  & ---                & 32                 \\
IB weight $\beta$          & ---                & $5\times10^{-3}$   \\
Feature layer $L_{\mathrm{feat}}$
                           & ---                & 17                 \\
Bottleneck hidden dimension & ---               & 512                \\
\bottomrule
\end{tabular}
\end{minipage}
\end{table*}

\FloatBarrier

\subsection{Visualization Analysis}
\label{sec:attn_vis}
\begin{figure}[ht]
    \centering
    \includegraphics[page=1,width=0.8\linewidth]{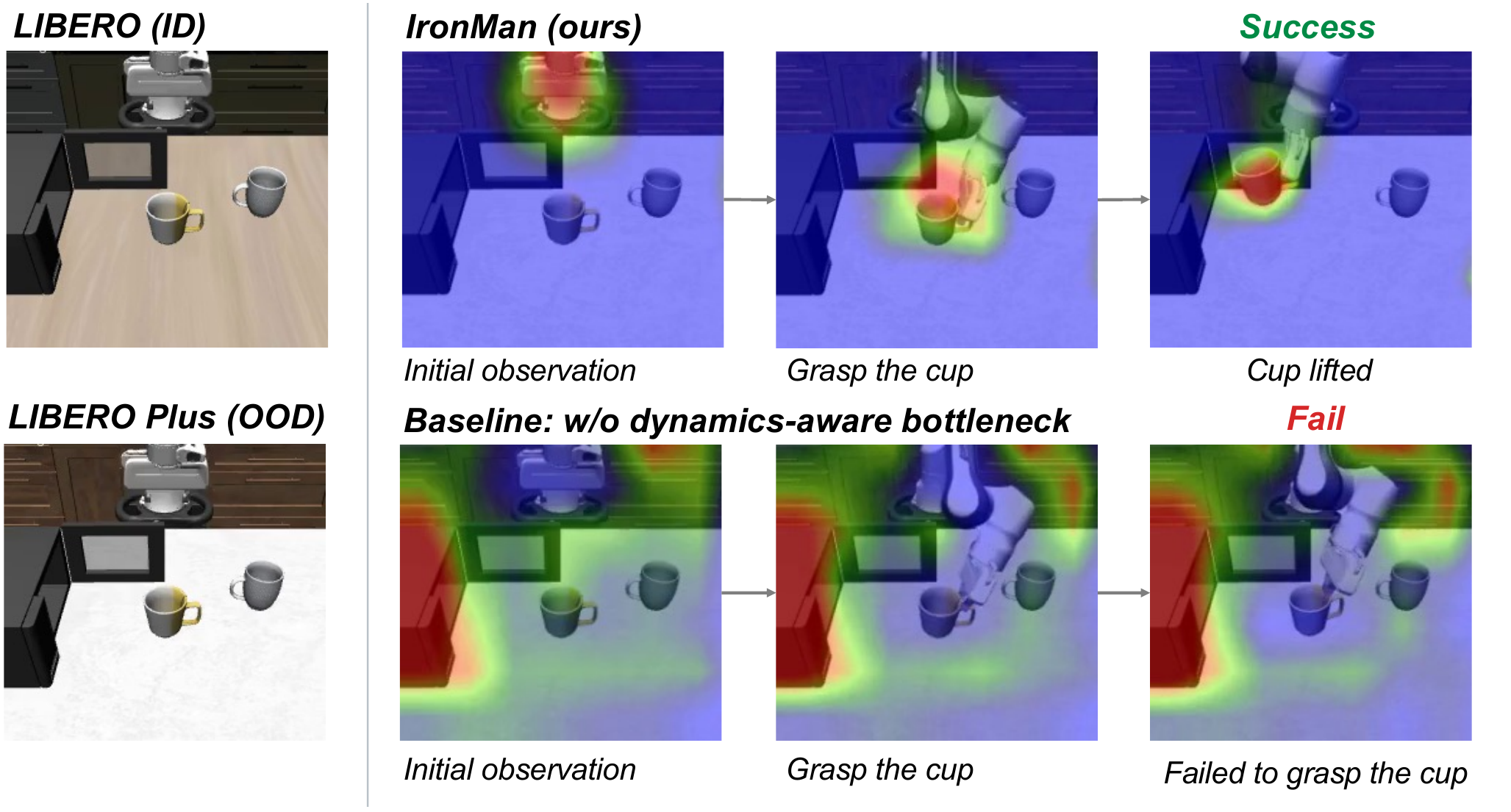}
    \caption{Attention visualization for grasping and lifting the cup.}
    \label{fig:attn_vis_all_1}
\end{figure}

\begin{figure}[ht]
    \centering
    \includegraphics[page=2,width=0.8\linewidth]{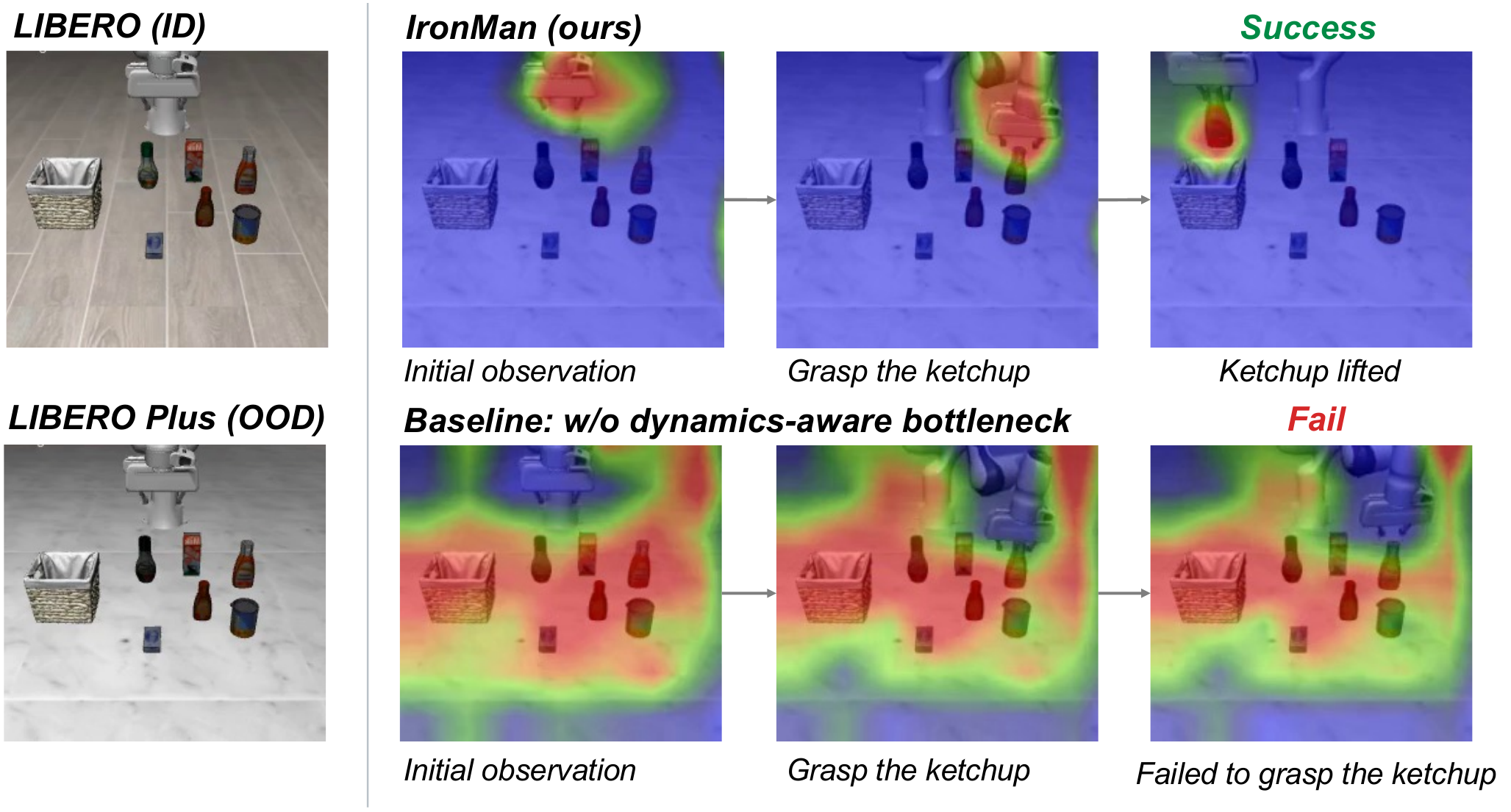}
    \caption{Attention visualization for grasping and lifting the ketchup bottle.}
    \label{fig:attn_vis_all_2}
\end{figure}

\begin{figure}[ht]
    \centering
    \includegraphics[page=3,width=0.8\linewidth]{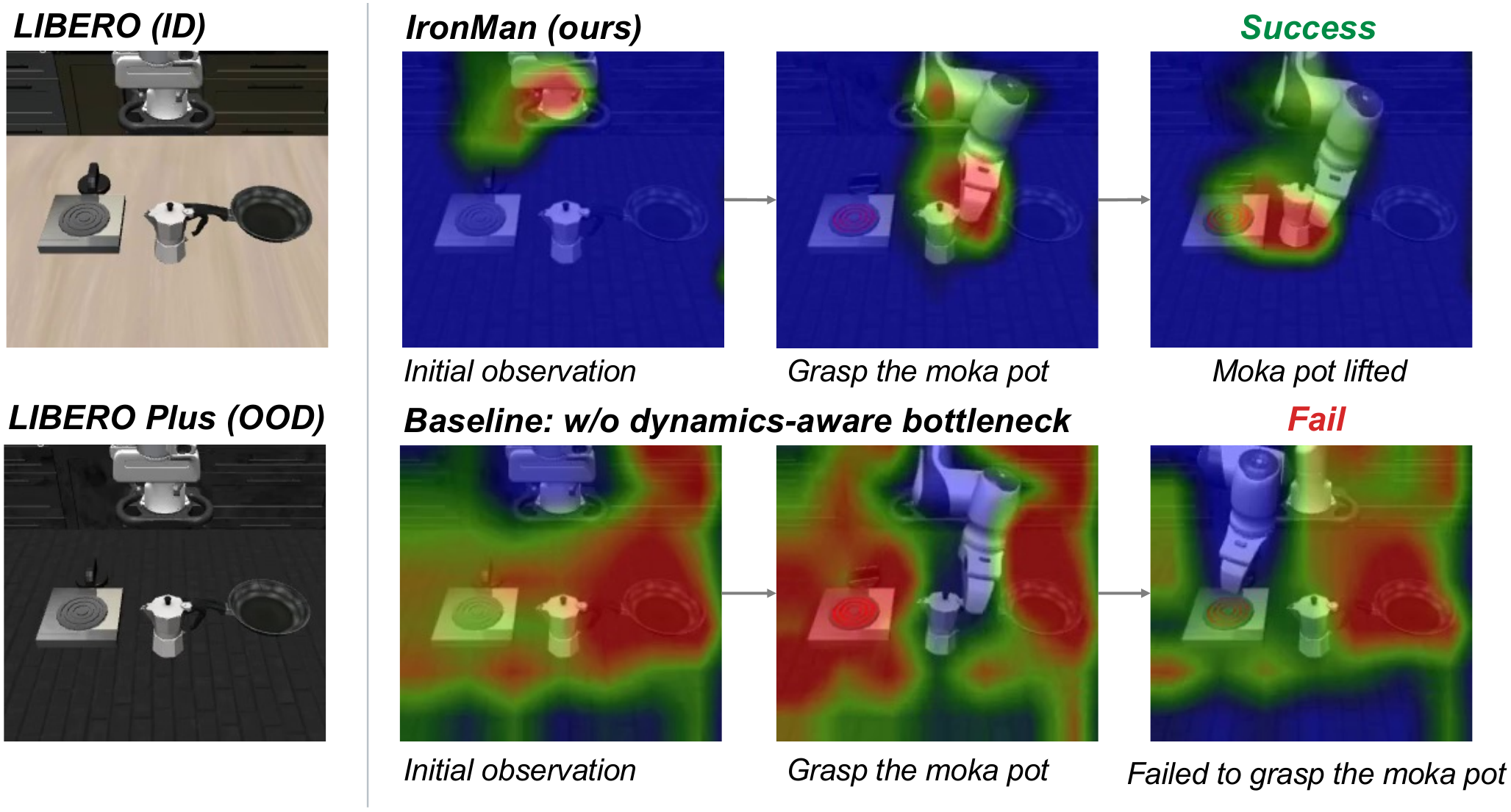}
    \caption{Attention visualization for grasping and lifting the moka pot.}
    \label{fig:attn_vis_all_3}
\end{figure}

\begin{figure}[ht]
    \centering
    \includegraphics[page=4,width=0.8\linewidth]{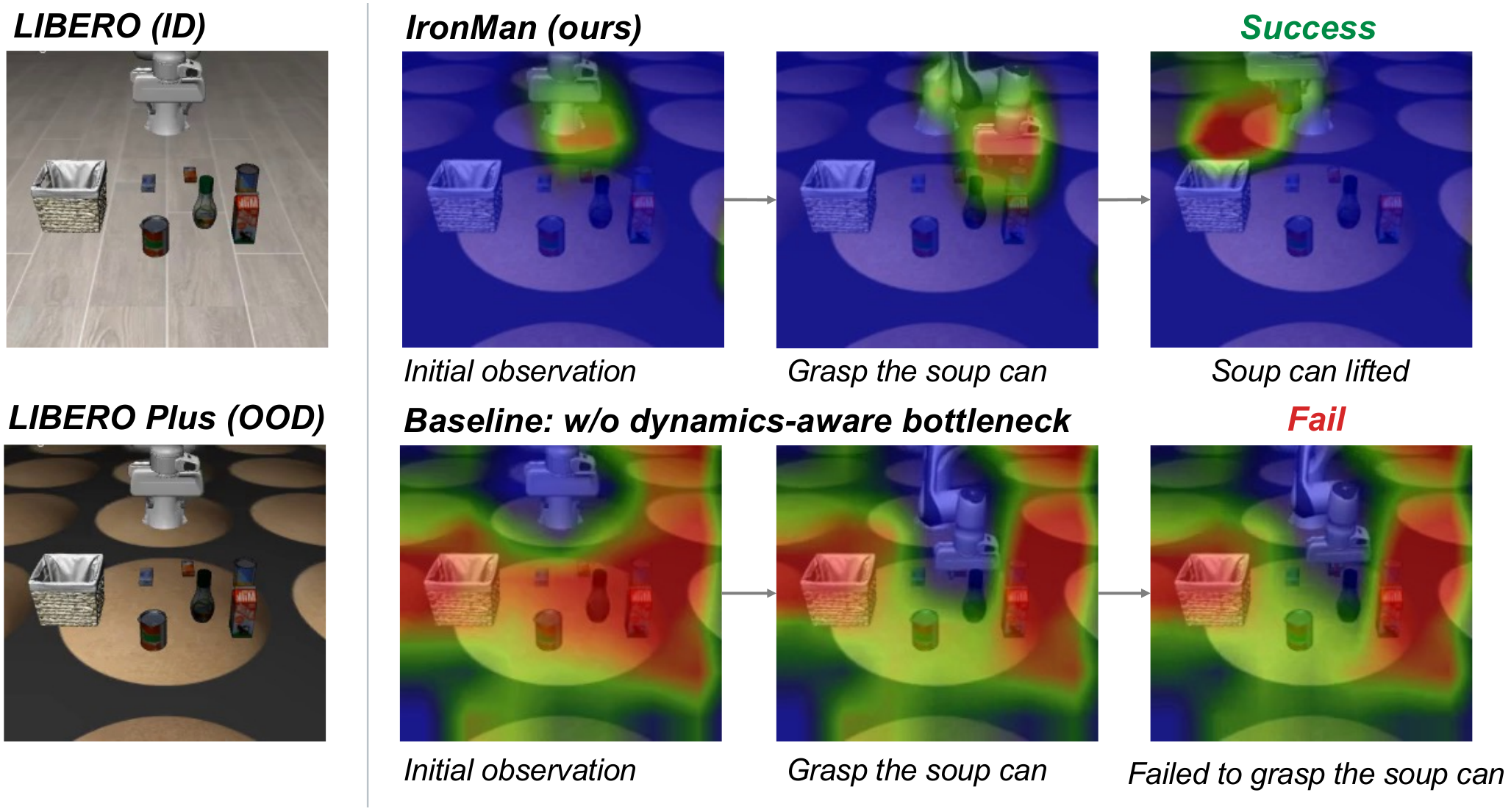}
    \caption{Attention visualization for grasping and lifting the soup can.}
    \label{fig:attn_vis_all_4}
\end{figure}

Figures~\ref{fig:attn_vis_all_1}--\ref{fig:attn_vis_all_4} extend the visualization analysis in \Figref{fig:sim_vis_main} with four examples involving a cup, a ketchup bottle, a moka pot, and a soup can. These examples cover different object arrangements and background shifts from LIBERO to LIBERO Plus, including changes in surface texture, color, and pattern. Across all four examples, IronMan exhibits more spatially selective attention than the baseline without the dynamics-aware bottleneck. Its attention is initially concentrated around the gripper and shifts toward the gripper--object interaction region as the robot approaches and grasps the target. During lifting, the high-attention region remains localized around the gripper and manipulated object, indicating that the spatial focus evolves with the manipulation rather than remaining fixed on background structures.

In contrast, the baseline exhibits broad attention over surrounding surfaces and non-target objects throughout the illustrated sequences. This difference is particularly visible in the cluttered ketchup-bottle and soup-can examples, where the baseline highlights large regions containing multiple objects and the basket, while IronMan maintains a more localized focus around the ongoing interaction. A similar pattern appears in the cup and moka-pot examples despite their different scene layouts: substantial baseline attention falls on surrounding structures or nearby objects, whereas IronMan emphasizes the region involved in grasping. In all four displayed rollouts, IronMan successfully grasps and lifts the target, while the baseline fails to grasp it. Together, these observations provide additional qualitative evidence consistent with the intended role of the dynamics-aware bottleneck: retaining manipulation-relevant information while reducing sensitivity to task-irrelevant visual variation.


\FloatBarrier

\subsection{Real-World Experiments Details}
\subsubsection{Task Demonstrations}
\label{sec:real_task_demo}
\begin{figure}[ht]
    \centering
    \includegraphics[page=1,width=\linewidth]{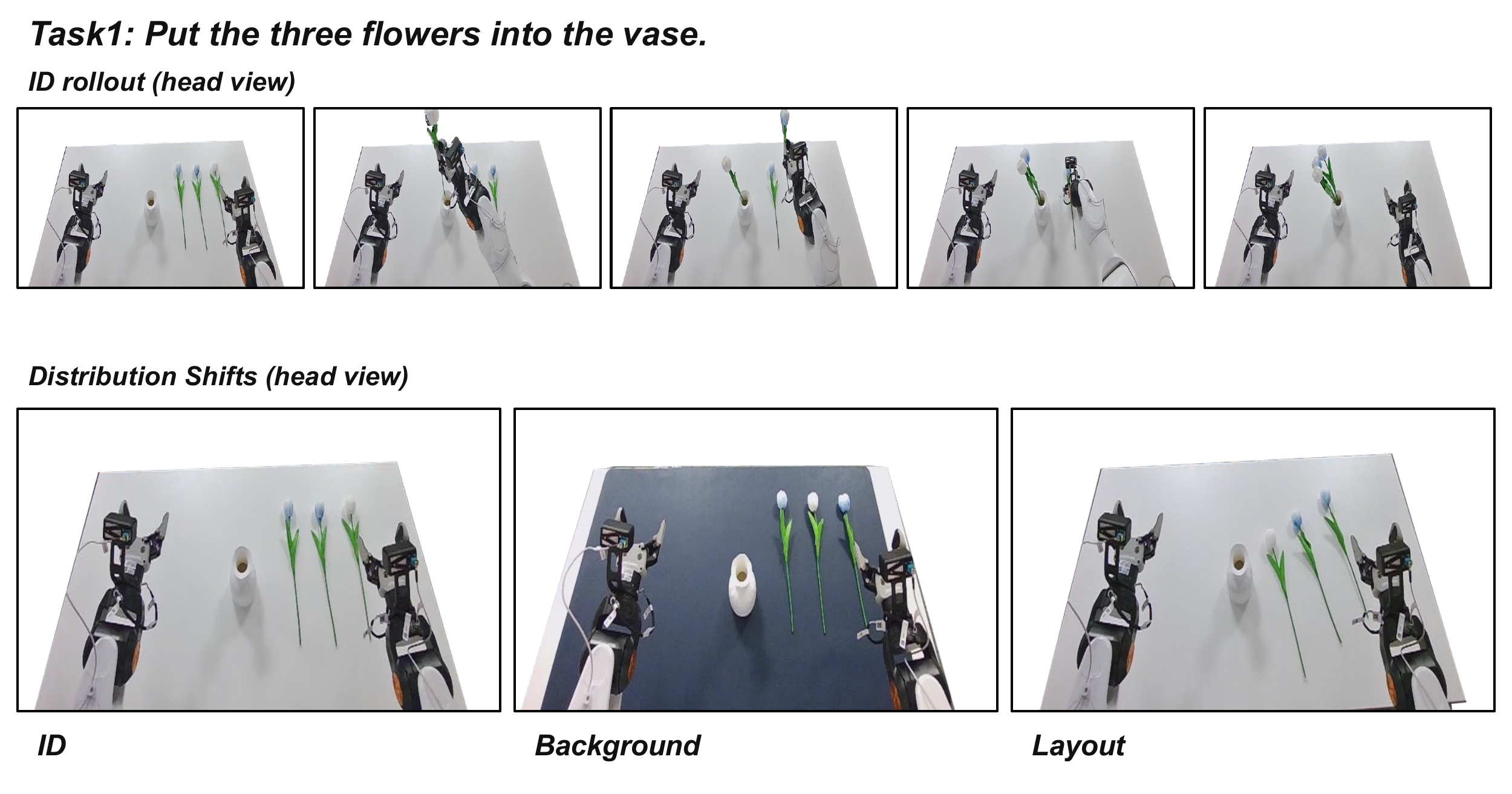}
    \includegraphics[page=2,width=\linewidth]{figures/real_world_examples.pdf}
    \includegraphics[page=3,width=\linewidth]{figures/real_world_examples.pdf}
    \caption{Real-world task demonstrations.}
    \label{fig:real_world_examples}
\end{figure}

As shown in~\Figref{fig:real_world_examples}, we evaluate our method on three real-world robotic manipulation tasks across two robbot platforms. The first two tasks use an AGIBOT G2 robot equipped with Omini Picker grippers, while the third uses a ROBOTERA M7 robot equipped with XHAND 1 dexterous hands. Both robots are equipped with three cameras: one head-mounted camera and two wrist-mounted cameras, one on each wrist. These cameras provide complementary views of the workspace and the regions around the end effectors. In Task 1, the robot must pick up three flowers and place them into a vase, requiring it to grasp each flower and align it with the vase opening before insertion. In Task 2, the robot must pick up fruits from the table and place them onto a plate, requiring object grasping and controlled placement within the target area. In Task 3, the robot must grasp parcels and place them on a shelf, requiring stable grasping during transport and accurate placement on the shelf.

Each task is evaluated under three settings: in-distribution (ID), Background, and Layout. The ID setting follows the training distribution and measures performance under familiar scene conditions. The Background setting introduces an out-of-distribution (OOD) shift by changing the tabletop background, testing robustness to changes in visual appearance. The Layout setting introduces a second OOD shift by changing the initial positions and orientations of the objects, testing generalization to unfamiliar spatial arrangements. These two OOD settings evaluate the effects of visual background changes and object layout changes, respectively.

We conduct 10 trials for each task under each setting, yielding 30 trials per task and 90 trials across the three tasks. Each trial is scored out of 10 points according to the task-specific scoring criteria detailed below.

\subsubsection{Score Protocol}
\label{app:score}
\Tabref{tab:real_world_scoring} shows the score protocol. We score each trial according to the highest sequential completion
stage achieved.
The flower task follows the order of the first, second, and third
flowers. The fruit task follows the order of banana, apple, lemon,
and pear. The parcel task requires placing the first small parcel
on the second shelf level, followed by the second small parcel on
the same level, and finally the large parcel on the first shelf level.
Each score level requires completion of all preceding stages. The same scoring criteria apply to the ID, Background, and Layout settings. We conduct 10 trials for each task under each setting and report normalized task scores on a scale from 0 to 100.

\begin{table}[ht]
    \footnotesize
    \centering
    \caption{Sequential scoring criteria for the three real-world tasks. Entries are raw per-trial scores on a 0--10 scale; reported task scores are normalized to a 0--100 scale.}
    \label{tab:real_world_scoring}
    \begin{tabularx}{0.8\linewidth}{
    @{} l c >{\raggedright\arraybackslash}X @{}
}
        \toprule
        Task & Score & Completion criterion \\
        \midrule

        \textbf{insert flowers}
        & 0    & The first flower is not placed in the vase. \\
        & 3.33 & The first flower is placed in the vase. \\
        & 6.67 & The second flower is placed in the vase. \\
        & 10   & The third flower is placed in the vase. \\

        \midrule
        \textbf{arrange fruits}

        & 0   & The banana is not placed on the plate. \\
        & 2.5 & The banana is placed on the plate. \\
        & 5   & The apple is placed on the plate. \\
        & 7.5 & The lemon is placed on the plate. \\
        & 10  & The pear is placed on the plate. \\

        \midrule

        \textbf{organize parcels}
        & 0    & No parcel is placed on shelf level 2. \\
        & 3.33 & The right small parcel is placed on shelf level 2. \\
        & 6.67 & The left small parcel is placed on shelf level 2. \\
        & 10   & The large parcel is placed on shelf level 1. \\

        \bottomrule
    \end{tabularx}

    \smallskip
\end{table}

\FloatBarrier

\subsection{Additional Experiments}
\subsubsection{Intervention Experiments}
\label{app:intervention}

We evaluate both models’ sensitivity to visual perturbations on one task from each of four LIBERO suites—Spatial, Object, Goal, and LIBERO-Long—with a fixed initial state per task. Each model generates and executes a 32-step action chunk open-loop under clean conditions, yielding eight reference trajectories. For each task, we sample 10 configurations without replacement from each of four LIBERO-plus perturbation categories—Camera Viewpoints, Light Conditions, Background Textures, and Sensor Noise—using a fixed random seed, yielding 320 perturbed trajectories across both models. All rollouts use the same task-specific physical initial state, instruction, initial proprioceptive input, and physical parameters; both models receive identical images for corresponding conditions, with inference seeds fixed. We measure the mean absolute difference (L1 error) relative to the same model’s clean reference for the same task.

\begin{figure}[htbp]
    \centering
    \includegraphics[width=\linewidth]{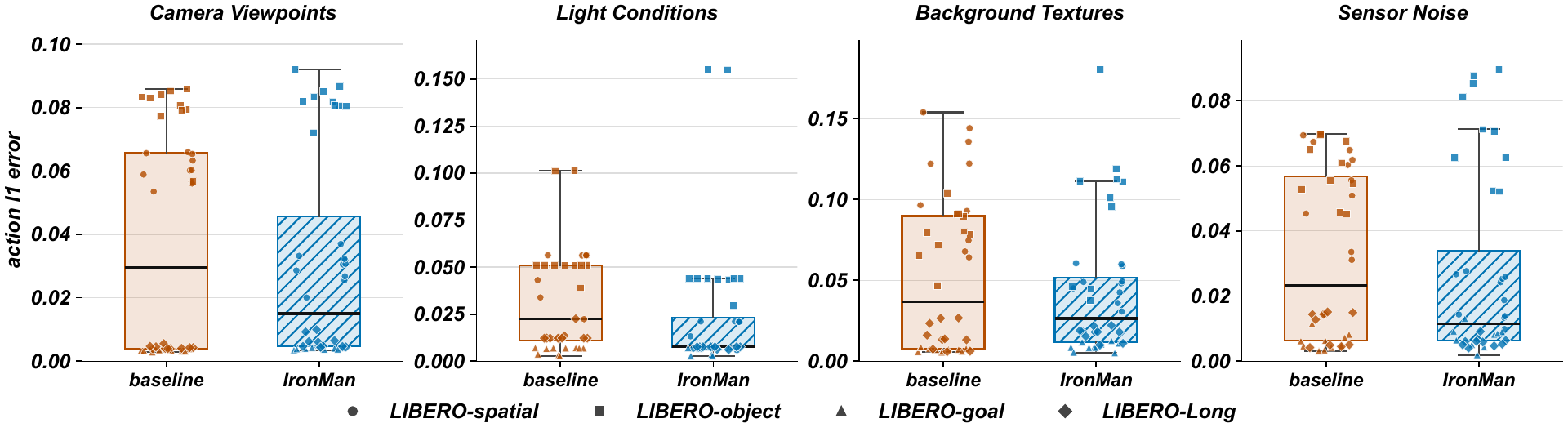}
    \caption{Action prediction error to visual perturbations. Box plots show the L1 error relative to each model's clean reference under four categories of visual perturbation, with 40 values per category and model across four LIBERO tasks and 10 perturbation configurations per task.}
    \label{fig:action_l1_boxplot}
\end{figure}

We show the results in ~\Figref{fig:action_l1_boxplot}. Across all four visual perturbation categories, IronMan exhibits lower median action L1 deviations and narrower interquartile ranges than the baseline. The reductions are particularly pronounced under changes in camera viewpoint, lighting, and sensor noise, while background texture perturbations yield a smaller reduction in the median. These results suggest that IronMan reduces both typical action deviations and their variability across the evaluated task–configuration pairs. Overall, IronMan improves action consistency under visual perturbations.

\end{document}